\documentclass[11pt]{article}

\usepackage[preprint]{acl}
\usepackage{times}
\usepackage{latexsym}
\usepackage[utf8]{inputenc}
\usepackage{inconsolata}

\title{\mytitle\vspace{-3mm}}

\author{Jordis Emilia Herrmann,  Aswath Mandakath Gopinath, Mikael Norrlof, Mark Niklas Müller \\
  Affiliation / Address line 1 \\
  Affiliation / Address line 2 \\
  Affiliation / Address line 3 \\
  \texttt{email@domain} \\\And
  Second Author \\
  Affiliation / Address line 1 \\
  Affiliation / Address line 2 \\
  Affiliation / Address line 3 \\
  \texttt{email@domain} \\}

\author{Jordis Emilia Herrmann\footnotemark[2]$^{1}$,  Aswath Mandakath Gopinath $^{1,2}$ \\ \textbf{Mikael Norrlof $^{1}$, Mark Niklas Müller $^{3,4}$}\vspace{2mm}\\ 
\hfil$^{1}$ ABB Robotics \hspace{2em} $^{2}$ Linköping University\hspace{2em} $^{3}$ LogicStar.ai \hspace{2em}$^{4}$ ETH Zurich\hfil
}

\usepackage[utf8]{inputenc} %

\usepackage{graphicx}

\usepackage[T1]{fontenc}

\IfFileExists{headers/config/showoverfull.config}{
	\overfullrule=1cm
}{
}

\usepackage{marginnote}

\usepackage[backgroundcolor=none,linecolor=red,textsize=footnotesize]{todonotes}

\usepackage{etoolbox}

\newbool{includeappendix}
\setbool{includeappendix}{true} %
\IfFileExists{headers/config/noappendix.config}{
	\setbool{includeappendix}{false}
}{}

\newif\ifincludeappendixx
\ifbool{includeappendix}{
	\includeappendixxtrue
}{
	\includeappendixxfalse
}

\usepackage{xr} %
\usepackage{filecontents}

\ifbool{includeappendix}{}{
	\input{appendix-labels-loader}

	\externaldocument{appendix-labels}
}

\usepackage{listings}

\usepackage{textcomp}

\usepackage{xcolor}

\usepackage[scaled=0.8]{beramono}

\definecolor{ckeyword}{HTML}{7F0055}
\definecolor{ccomment}{HTML}{3F7F5F}
\definecolor{cstring}{HTML}{2A0099}

\lstdefinestyle{numbers}{
	numbers=left,
	framexleftmargin=20pt,
	numberstyle=\tiny,
	firstnumber=auto,
	numbersep=1em,
	xleftmargin=2em
}

\lstdefinestyle{layout}{
	frame=none,
	captionpos=b,
}

\lstdefinestyle{comment-style}{
	morecomment=[l]//,
	morecomment=[s]{/*}{*/},
	commentstyle={\color{ccomment}\itshape},
}

\lstdefinestyle{string-style}{
	morestring=[b]",%
	morestring=[b]',%
	stringstyle={\color{cstring}},
	showstringspaces=false,%
}

\lstdefinestyle{keyword-style}{
	keywordstyle={\ttfamily\bfseries},
	morekeywords={
		function,
		constructor,
		int,
		bool,
		return,
		returns,
		uint
	},
	morekeywords = [2]{},
	keywordstyle = [2]{\text},
	sensitive=true,
}

\lstdefinestyle{input-encoding}{
	inputencoding=utf8,
	extendedchars=true,
	literate=
	{ℝ}{$\reals$}1%
	{→}{$\rightarrow$}1%
	{α}{$\alpha$}1%
	{β}{$\beta$}1%
	{λ}{$\lambda$}1%
	{θ}{$\theta$}1%
	{ϕ}{$\phi$}1%
}

\lstdefinestyle{escaping}{
	moredelim={**[is][\color{blue}]{\%}{\%}},
	escapechar=|,
	mathescape=true
}

\lstdefinestyle{default-style}{
	basicstyle=\fontencoding{T1}\ttfamily\footnotesize,
	style=numbers,
	style=layout,
	style=comment-style,
	style=string-style,
	style=keyword-style,
	style=input-encoding,
	style=escaping,
	tabsize=2,
	upquote=true
}

\lstdefinelanguage{BASIC}{
	language=C++,
	style=default-style
}[keywords,comments,strings]%

\usepackage{tikz}

\usetikzlibrary{arrows}
\usetikzlibrary{automata}
\usetikzlibrary{calc}
\usetikzlibrary{backgrounds}
\usetikzlibrary{decorations.markings}
\usetikzlibrary{decorations.pathmorphing}
\usetikzlibrary{decorations.pathreplacing}
\usetikzlibrary{fit}
\usetikzlibrary{patterns}
\usetikzlibrary{positioning}
\usetikzlibrary{shadows}
\usetikzlibrary{shapes}
\usetikzlibrary{shapes.geometric}
\usetikzlibrary{arrows.meta, calc}

\usepackage{amsmath,amsfonts,bm}

\def\1{\bm{1}}

\def\mA{{\bm{A}}}
\def\mB{{\bm{B}}}

\def\mW{{\bm{W}}}

\DeclareMathAlphabet{\mathsfit}{\encodingdefault}{\sfdefault}{m}{sl}
\SetMathAlphabet{\mathsfit}{bold}{\encodingdefault}{\sfdefault}{bx}{n}

\newcommand{\R}{\mathbb{R}}

\usepackage{microtype}
\usepackage{graphicx}
\usepackage{subfigure}
\usepackage{booktabs} %
\usepackage{hyperref}
\usepackage[symbol]{footmisc}

\usepackage{amsmath}
\usepackage{amssymb}
\usepackage{mathtools}
\usepackage{amsthm}
\usepackage{xspace}
\usepackage{siunitx}
\usepackage{multirow}
\usepackage{tcolorbox}

\newcolumntype{x}[2]{S[table-format=#1.#2,table-auto-round]}

\usepackage[capitalize,noabbrev]{cleveref}

\theoremstyle{plain}

\theoremstyle{definition}

\theoremstyle{remark}

\newcommand{\bench}{\textsc{SysDiagBench}\xspace}
\newcommand{\mytitle}{Diagnosing Robotics Systems Issues with Large Language Models}

\newcommand{\lora}{\textsc{LoRA}\xspace}
\newcommand{\qlora}{\textsc{QLoRA}\xspace}

\newcommand{\mistral}{\textsc{Mistral-Lite-7B}\xspace}
\newcommand{\mixtral}{\textsc{Mixtral-8x7B}\xspace}
\newcommand{\gptf}{\textsc{GPT-4}\xspace}

\usepackage[capitalize]{cleveref}

\makeatletter
\newcommand\theHALG@line{\thealgorithm.\arabic{ALG@line}}
\makeatother

\crefrangeformat{section}{\S#3#1#4\crefrangeconjunction\S#5#2#6}

\crefmultiformat{section}{\S#2#1#3}{\crefpairconjunction\S#2#1#3}{\crefmiddleconjunction\S#2#1#3}{\creflastconjunction\S#2#1#3}

\newcommand{\crefrangeconjunction}{--}

\crefname{listing}{Lst.}{listings}
\crefname{line}{Lin.}{Lin.}

\newcommand{\appref}[1]{%
	\ifbool{includeappendix}{\cref{#1}}{the appendix}%
}
\newcommand{\Appref}[1]{%
	\ifbool{includeappendix}{\cref{#1}}{The appendix}%
}

\begin{document}
\maketitle

\begin{abstract}
Quickly resolving issues reported in industrial applications is crucial to minimize economic impact. However, the required data analysis makes diagnosing the underlying root causes a challenging and time-consuming task, even for experts. In contrast, large language models (LLMs) excel at analyzing large amounts of data. Indeed, prior work in AI-Ops demonstrates their effectiveness in analyzing IT systems. Here, we extend this work to the challenging and largely unexplored domain of robotics systems. To this end, we create \bench, a proprietary system diagnostics benchmark for robotics, containing over $2\,500$ reported issues. We leverage \bench to investigate the performance of LLMs for root cause analysis, considering a range of model sizes and adaptation techniques. Our results show that \qlora finetuning can be sufficient to let a $7$B-parameter model outperform \gptf in terms of diagnostic accuracy while being significantly more cost-effective. We validate our LLM-as-a-judge results with a human expert study and find that our best model achieves similar approval ratings as our reference labels.

\end{abstract}
\vspace{-3mm}
\section{Introduction}
\label{sec:intro}
\vspace{-1mm}
\footnotetext[2]{Correspondence author: jordis.herrmann@se.abb.com}
\let\svthefootnote\thefootnote

Identifying the root cause for issues with complex industrial systems is a time-critical task but challenging and time-consuming for human experts as they struggle with the required analysis of large amounts of log data which large language models (LLMs) excel at. Indeed, there is substantial work in AI-Ops exploring automated diagnostics for IT systems \citep{DiazdeArcayaTZMA24,ZhaoxueLZGYL21}. However, the challenging domain of robotics systems remains largely unexplored.

\paragraph{This Work: Automated Diagnostics for Robotics Systems} To address this challenge, we create \bench, a proprietary benchmark for diagnosing root causes of complex robotics systems failures, containing over $2\,500$ real-world issues. In particular, each instance corresponds to a support ticket, containing an issue description, a set of log files, communications with the support engineers, a reference root cause extracted from expert discussions, and the ultimate issue resolution. The goal in \bench is to predict the root cause underlying the reported issue, given only the information available at the creation of the ticket. 

\paragraph{LLM-Based Diagnostics} We leverage \bench to investigate multiple LLM-based diagnostic approaches in the robotics setting, using both LLM-as-a-judge \citep{ZhengC00WZL0LXZ23} and human experts for evaluation. In particular, we consider a range of model sizes and adaptation techniques from zero-shot prompting to full finetuning, to assess their cost-performance trade-off. Interestingly, we observe that even \qlora \citep{DettmersPHZ23} can be sufficient to let a $7$B-parameter model outperform $\gptf$ in terms of diagnostic accuracy while being significantly more cost-effective. Validating our results in an expert study, we find that LLM-as-a-judge scores correlate well with human expert ratings, with our reference labels matching the experts' analysis in over half the cases and our best model achieving similar approval ratings as these reference labels.

\paragraph{Key Contributions} Our key contributions are:
\vspace{-2mm}
\begin{itemize}
    \renewcommand{\itemsep}{0em}
    \item We create \bench, a proprietary benchmark for automated root cause analysis of robotics systems, based on thousands of real-world issues (\cref{sec:benchmark}).
    \item We propose a range of LLM-based diagnostic tools (\cref{sec:method}).
     \item We leverage \bench to analyze these techniques and identify the most effective and efficient strategies (\cref{sec:experiments}).
    \item We validate the effectiveness of our approach using a human expert study (\cref{sec:human_study}).
\end{itemize}

\section{Related Work}
\label{sec:related}
\vspace{-1mm}

\paragraph{AI-Ops} leverages machine learning (ML) in IT operations \citep{DiazdeArcayaTZMA24} to analyze large amounts of semi-structured data such as logs and traces \citep{ZhaoxueLZGYL21} with the goal of discovering anomalies and their root causes. As many traditional ML methods require structured data, AI-Ops long focused on developing methods enhancing \citep{YuanZPZS12,ZhaoRLSYZ17} and parsing \citep{HeZZL17,MessaoudiPBBS18} log files, using well-established methods such as SVMs \citep{ZhangS08,ZuoWMHP20}, simple clustering techniques \citep{ZhaoCBRBMC19,LouFYXL10}, and decision tree (ensembles) \citep{ChenZLJB04} for the actual analysis. 

\vspace{-2mm}
\paragraph{LLM-based Approaches}
As LLMs can directly process the semi-structured log data, they have recently gained popularity in the field \citep{shao2022log,ChenL22b,LeeKK23a, abs-2102-11570}. As a representative example, \citet{GuptaKKBAM23} use an encoder architecture, pre-trained on a large amount of log data, to compute embeddings for further analysis. In contrast to these methods, we propose to directly predict root causes from log data.

\vspace{-1mm}
\section{\bench: A Benchmark for Robotics System Diagnostics}
\label{sec:benchmark}

\bench is a proprietary system diagnostics benchmark focusing on root causes (RC) prediction for real-world robotics issues, constructed from a decade of industry data. 
Concretely, each \bench instance corresponds to a support ticket and contains a detailed problem description, a set of log files from the affected system, and a reference root cause description. Below, we first describe the information contained in a ticket and then the process of constructing \bench. Unfortunately, the underlying data cannot be published at this point due to privacy concerns.

\paragraph{Support Tickets} A ticket is created when a reported issue cannot be resolved by the service support engineers and as a result needs to be escalated to the product development team. Every ticket contains metadata on the affected system (e.g. the robot and application type), a detailed problem description, and a system diagnostic file capturing the system state after the issue occurred. For \bench, we consider three log files contained in the system diagnostic that are commonly analyzed by experts when investigating a ticket.
The \texttt{elog} logs all error, warning, and information events that occur during the operation of the system. %
The \texttt{print-spool} logs all outputs that are written to the console during the operation of the system.
The \texttt{startup} logs all events that occur during the startup of the system. %
All three log types contain a significant amount of data at average lengths of $37$k, $23$k, and $8.1$k tokens, respectively.

\paragraph{Historic Tickets} which were already resolved successfully additionally contain the discussion among the experts working on the issue, communication with the support engineers, and final resolution of the issue. However, as even these historic tickets generally do not contain a description of the root cause (RC), we need to extract it from the available information to obtain a reference label.%

\begin{figure}
    \centering
    \resizebox{\linewidth}{!}{\input{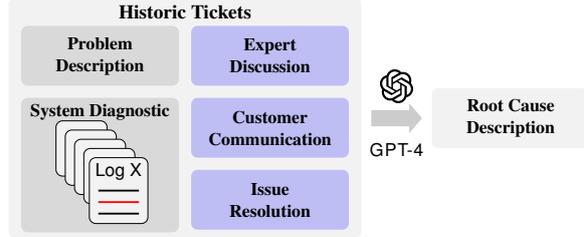}}
    \vspace{-7mm}
    \caption{Visualization of the label extraction process for historic tickets based on querying a strong LLM. Note that during inference time only the grey, but not the blue, boxes are available.}
    \label{fig:label_extraction}
    \vspace{-5mm}
\end{figure}

\subsection{Benchmark Construction}
\label{sec:benchmark_construction}

To create \bench, we collected over $12\,000$ historic tickets and filtered out those that do not contain a system diagnostic with an \texttt{elog}, \texttt{pspool}, and \texttt{startup} file, leaving us with $2\,585$, split into a training, validation, and test set corresponding to $75\%$, $5\%$, and $20\%$, respectively.

\paragraph{Root Cause Extraction}
To extract a concise root cause description from a historic ticket, we leverage a strong LLM (\gptf) (illustrated in \cref{fig:label_extraction}). Concretely, we query the LLM with the problem description, expert discussion, support engineer communication, and final resolution using a chain-of-thought (CoT) prompt \citep{Wei0SBIXCLZ22}, instructing the model to carefully analyze all provided information before describing the root cause (see \cref{sec:app_prompt_extraction} for more details). We highlight that the information used to create these labels is not available when a ticket is created and can thus not be used to predict the RC at inference time. Finally, we validate the quality of the extracted root causes in a human study in \cref{sec:human_study}.

\subsection{Evaluation Metrics}
Evaluating root cause correctness is inherently challenging, as descriptions of the same, correct root cause can be highly diverse, making similarity measures such as the ROUGE score \citep{lin2004rouge} unsuitable. Further, there is frequently a trade-off between specificity and correctness, i.e., generic descriptions can be correct yet unhelpful, while very precise ones may get minor details wrong while still being overall very helpful. We thus adopt an LLM-as-a-judge evaluation \citep{ZhengC00WZL0LXZ23} asking a model to judge the similarity between the predicted and the reference RC on a scale of $1$ to $10$ (see \cref{sec:app_prompt_sim} for details). We report the mean similarity score ($MSS$) with respect to the reference labels as our primary evaluation metric and validate it against human experts in \cref{sec:human_study}.

\vspace{-1mm}
\section{LLM-Based Systems Diagnostic}
\label{sec:method}
\vspace{-1mm}
In this section, we describe the system diagnostic approaches we evaluated on \bench. 

\vspace{-1mm}
\subsection{Input Preprocessing and Prompting}
\begin{figure}
    \centering
    \includegraphics[width=0.82\linewidth]{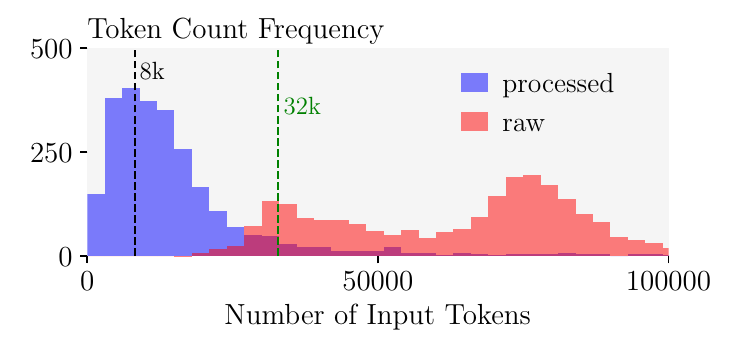}
    \vspace{-6mm}
    \caption{Token count distribution of processed finetuning inputs (blue) and corresponding raw logs (red).}
    \label{fig:input_lengths}
    \vspace{-5mm}
\end{figure}

\vspace{-1mm}
For both training and inference, we preprocess all log files by removing timestamps, dates, and sequence numbers. 
We further remove all consecutive duplicate lines and filter the \texttt{elog} to only include error and warning but not information events, as these are most likely to be relevant for diagnosing the root cause. 
As both the \texttt{print-spool}  and \texttt{startup} frequently contain tens of thousands of lines, we only retain the $10$ lines before and after each error or warning event in the \texttt{elog} file. Finally, while the original \texttt{elog} file contains only integer error IDs, we map these to human-readable error descriptions using a lookup table. This preprocessing reduces the mean total token count of the log files per ticket from $68$k to $16$k tokens, with the distribution change illustrated in \cref{fig:input_lengths}.

The LLM input is now constructed by combining a detailed CoT instruction \citep{Wei0SBIXCLZ22} with a context of the preprocessed log files and the issue description (see \cref{sec:app_prompt_prediction} for more details). Despite our preprocessing, the resulting inputs frequently exceed 8k ($68\%$) and even $32$k ($9\%$) tokens (see \cref{fig:input_lengths}), requiring models with large context capabilities for processing.

\subsection{Training for System Diagnostics}
\vspace{-1mm}
\label{sec:adaptation_methods}
While modern LLMs have impressive zero-shot capabilities \citep{KojimaGRMI22}, adapting them to specific tasks \citep{zhao2024lora} can improve their performance significantly.
However, the long input lengths make in-context learning, e.g., via few-shot prompting, unpractical for system diagnostics. We, thus, consider three adaptation techniques, full finetuning (FFT), \lora \citep{HuSWALWWC22}, and \qlora \citep{DettmersPHZ23}, with different performance-cost trade-offs (see \cref{sec:app_background}).

\vspace{-1mm}
\section{Experimental Evaluation}
\label{sec:experiments}
\vspace{-1mm}

Below, we discuss the experimental setup and key results. For more details on setup and extended results, please see \cref{sec:app_setup,sec:app_experiments}, respectively.

\vspace{-2mm}
\paragraph{Experimental Setup}
We consider \mistral \citep{MistralLite_2023}, \mixtral \citep{Mixtral2024}, and \gptf \citep{GPT4}, using \citet{Axolotl} for finetuning on $2$ to $8$ NVIDIA A100s. We train for 3 epochs and unless indicated otherwise,  use rank $r=32$ for (Q)\lora and NFloat4 + DQ (double quantization) for \qlora.

\begin{table}[h]
    \vspace{-2mm}
    \caption{Mean similarity score $MSS$ for different models and adaptation methods on \bench.}
    \vspace{-3mm}
    \label{tab:sys_diag}
    \centering
    \resizebox{0.93\linewidth}{!}{
    \begin{tabular}{l
                    x{2}{2}
                    @{\hskip 10mm}
                    x{2}{2}
                    x{2}{2}
                    x{2}{2}
        }
        \toprule
        Model & {Base} & {FFT} & {\lora} & {\qlora} \\
        \midrule
        \mistral & 2.39 & 2.94 & 3.07 & {$\;\,$}\textbf{3.27} \\ 
        \mixtral & 2.25 & 2.69 & 2.24 & 2.30 \\ 
        \gptf & {$\;\,$}\textbf{2.52} & {-} & {-} & {-} \\ 
        \bottomrule
    \end{tabular}
    }
    \vspace{-4mm}
\end{table}    

\paragraph{System Diagnostic Performance}
We compare the performance of different models and training methods in \cref{tab:sys_diag}. Interestingly, we find that the smaller \mistral outperforms \mixtral across all adaptation settings. We hypothesize this is because it was specifically trained for long context capabilities, crucial for analyzing long log files. While \gptf is the best-performing base model, we find that our finetuned models outperform it by a significant margin, with \qlora training yielding the best performance.

\begin{figure}[t]
    \centering
    \includegraphics[width=0.8\linewidth]{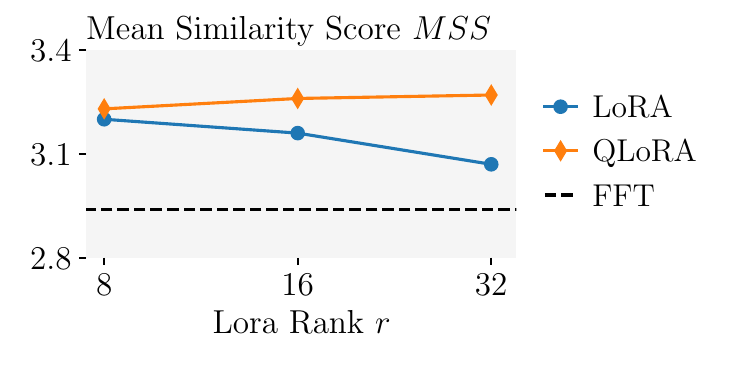}
    \vspace{-5mm}
    \caption{Mean similarity score of \mistral for \lora and \qlora training depending on rank $r$.}
    \vspace{-4mm}
    \label{fig:lora_rank}
\end{figure}

\vspace{-2mm}
\paragraph{Rank Deficiency as Regularization}
Observing that \lora and \qlora outperform full finetuning for \mistral (see \cref{tab:sys_diag}), we investigate the impact of the \lora rank $r$ on the model's performance, illustrating results in \cref{fig:lora_rank}. We find that while a smaller rank improves performance for \lora, there is only a minimal effect for \qlora. As both a reduced rank and quantization act as regularizers, we conclude that \qlora is already at optimal regularization, while \lora still benefits from further regularization.

\vspace{-1mm}
\section{Human Study}
\label{sec:human_study}
\vspace{-1mm}
We conduct a study with human experts to answer the following three questions: \emph{Q1:} Are the automatically extracted reference labels accurate? \emph{Q2:} Does our LLM-as-a-judge evaluation correlate well with human expert ratings? And \emph{Q3:} Are our best models helpful in root cause analysis? 

\paragraph{Study Setup} We ask $10$ experts to solve a subset of tickets, leading to $n\!=\!53$ answer sets. 
We provide the experts with all the tools they typically use to resolve issues and the full historic ticket data. We then ask them to describe the issue's RC and judge their confidence. Next, we let them assess our reference label, in terms of helpfulness (yes, no, maybe), correctness (on a scale from $1$ to $10$), and preference compared to their own answer. Where available, %
we let them assess their colleague's RC, in the same way. Finally, we let them rate the RCs generated by our four best models on the same scale. See \cref{sec:app_human_study} for more details.

\paragraph{Q1: Reference RC Quality} Asked directly, whether our reference RC was correct, experts agreed (yes or maybe) in $49\%$ of cases, saying it was as good as their own assessment in $55\%$ of cases, but only as good as their colleague's in $29\%$. Interestingly, they still assigned a higher or equal score to our reference RC ($6.0$ on average) than to their colleagues' RC ($7.9$ average) in $43\%$ of cases. Combined with experts only being highly (moderately) confident in their assessment $32\%$ ($58\%$) of the time, this suggests that diagnosing root causes is a particularly hard task, with our reference labels having high but not perfect quality.

\vspace{-1mm}
\paragraph{Q2: LLM-as-a-Judge Evaluation} %
While overall $MSS$ (LLM-as-a-judge) and mean expert ratings induce the same ranking, the $MSS$ evaluated on the samples considered by the experts ranks the fully finetuned $\mistral$ first instead of third. Comparing per-sample ratings, we find a correlation of $\rho = 0.20$, increased to $0.34$ on samples where our reference labels are considered correct by the experts. This matches the per-sample correlation of human expert scores of $\rho \!=\! 0.32$. 
These results suggest that our LLM-as-a-judge evaluation is a good proxy for a human expert judgment where we have reliable reference labels and allows for reliably comparing models.

\begin{figure}
    \centering
    \includegraphics[width=1.0\linewidth]{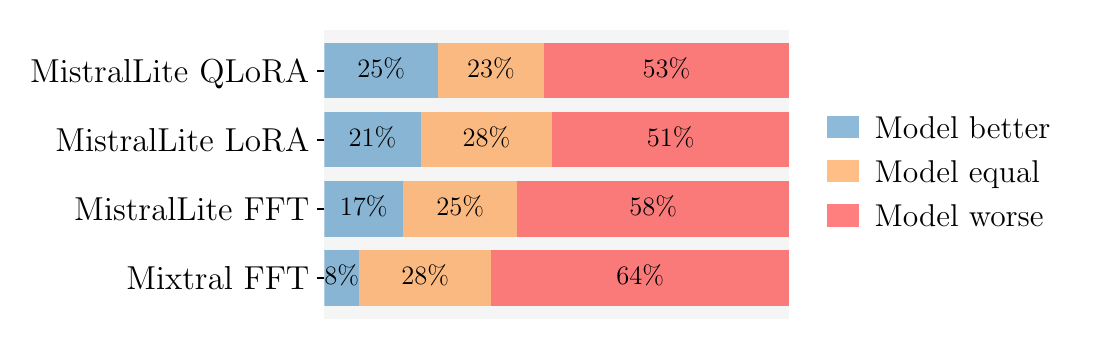}
    \vspace{-9mm}
    \caption{Frequency of human experts rating predicted RCs higher (blue), equal (orange), and lower (red) than our reference RC.}
    \label{fig:model_colleague_comparison}
    \vspace{-5mm}
\end{figure}

\vspace{-1mm}
\paragraph{Q3: Model Helpfulness} We compare the expert ratings of our reference and predicted RCs in \cref{fig:model_colleague_comparison} and observe that the RCs predicted by our best two models match the reference labels extracted by \gptf with hindsight knowledge in half the cases.  We further find that even when the \emph{reference} labels are considered incorrect, there are instances where the \emph{predicted} RCs are rated highly (see \cref{fig:rating_by_GT_correct} in \cref{sec:app_experiments}). For these instances, we found by manual inspection, that our models were able to leverage a deeper understanding of the log files to identify the underlying RC. These results suggest that our models can provide valuable insights and help experts in diagnosing the root causes of complex robotics systems issues.

\vspace{-1mm}
\section{Conclusion}
\vspace{-1mm}
We created \bench, a benchmark for robotics systems diagnostics, containing over $2\,500$ real-world issues. We leveraged \bench to investigate the performance of large language models (LLMs) for automated root cause analysis, considering a range of model sizes and adaptation techniques. Our results show that \qlora finetuning allows a $7$B-parameter model to outperform \gptf in terms of diagnostic accuracy while being significantly more cost-effective. We validated our results with an expert study and found that while both our reference label extraction and LLM-as-a-judge evaluation cannot replace human experts, our best models can provide valuable insights.

\section{Limitations}
\label{sec:limitations}
As no ground truth root cause annotations exist for (historic) tickets, we generated reference labels for \bench using a strong LLM to extract them from the rich data available for historic tickets. While the human expert study shows the generated labels to be as good as an expert’s analysis in over half the cases, they are not perfect. In particular, they are only considered correct (yes or maybe) $49\%$ of the time by the experts. Using multiple experts to annotate the same tickets could have improved the quality of the reference labels and thus both the performance of our finetuned models and the evaluation quality, but this was not feasible due to the significant effort required to annotate tickets and time constraints of available experts.

Further, even for correct reference labels, the LLM-as-a-judge evaluation is not perfect. While we achieve a high correlation of $\rho = 0.57$ between the $MSS$ and mean expert ratings when only considering samples with correct (yes or maybe) reference labels, the per-sample correlation remains at a moderate $\rho = 0.20$. However, even the inter-expert correlation of $\rho = 0.77$ for mean scores and $\rho = 0.32$ for per-sample correlation, remains far from perfect, highlighting again the difficulty of accurately assessing root causes. We conclude that while the LLM-as-a-judge evaluation is a valuable tool for comparing different models, it cannot substitute expert human judgment, especially for sample-level comparison. We note that for reliable sample-level comparison, a panel of experts would be needed.

Finally, we only consider two base models and a limited number of hyperparameters for our experiments due to both budget and time constraints. While we find for \mistral that \lora and \qlora finetuning perform exceptionally well, even outperforming full finetuning, this was not the case for \mixtral. While we hypothesize that this is due to \mistral's training specifically for long context retrieval tasks, a detailed study of this effect is out of scope here and left for future work.

\section{Ethical Considerations and Broader Impact}
\label{sec:ethical}
\paragraph{Ethical Considerations}
The dataset underlying \bench contains real-world support tickets from a robotics company, which may contain sensitive information about the company's products and customers, thus precluding its public release. To mitigate the risk of privacy breaches during internal use, we have anonymized all tickets by removing all personally identifying information fields. 

For our human expert study, we have recruited internal experts from the robotics company's product team, who regularly handle the support tickets constituting \bench. We have obtained informed consent from all participants and have anonymized the expert's analysis before sharing it with other experts for the inter-expert assessment. All experts were paid their regular wages during their participation in the study.

\paragraph{Broader Impact}
While we demonstrate the effectiveness of LLM-based systems for the automated diagnostics of robotics systems, we also highlight their limitations. In particular, we show that while LLMs can achieve moderately high performance and even match experts in some cases, they are unable to fully replace the expert's analysis. We thus expect that LLM-based tools will soon become useful aids for human experts, but not fully replace them in the foreseeable future, similar to many other domains.

\message{^^JLASTBODYPAGE \thepage^^J}

\clearpage
\bibliography{references}

\message{^^JLASTREFERENCESPAGE \thepage^^J}

\ifincludeappendixx
	\clearpage
	\newpage
	\appendix
	\onecolumn
	\section{Backgorund}
\label{sec:app_background}

\paragraph{Prompting} Once the remarkable zero-shot capabilities of LLMs had been demonstrated \citep{KojimaGRMI22}, a wide range of prompting schemes was proposed that aim to elicit higher quality answers from the same model by evoking a (more thorough) reasoning process \citep{0002WSLCNCZ23,YaoYZS00N23,XuYLZMZ23,ZhouSHWS0SCBLC23}. In particular, Chain-of-Thought (CoT) prompting \citep{Wei0SBIXCLZ22} instructs the model to "think step-by-step" when answering, which has been shown to improve performance on a wide range of tasks, with multiple follow-up works trading-off increased inference cost and better performance \citep{0002WSLCNCZ23,YaoYZS00N23}.

\paragraph{(Full) Finetuning} If zero-shot performance is unsatisfactory and labeled training data is available, one can continue training the model on the specific task at hand, a process known as finetuning. In particular, full finetuning refers to training the entire model on the new task.

\paragraph{\lora}
However, the huge size of modern LLMs makes GPU memory a bottleneck for training, with common optimizers like Adam \citep{KingmaB14} and AdamW \citep{LoshchilovH19} requiring three full precision values (the gradient, and its first and second moment) to be tracked for every parameter. To alleviate this issue, \lora \citep{HuSWALWWC22} proposes that instead of updating all parameters in a weight matrix $\mW \in \R^{n \times n}$ one only computes a low-rank update $\mA\mB$ where $\mA \in \R^{n \times k}$ and $\mB \in \R^{k \times n}$ with $k \ll n$. We thus obtain the updated weight matrix as $\mW' = \mW + \mA\mB$ and reduce the memory footprint of the optimizer from $O(n^2)$ to $O(nk)$. Finally, recent work has shown that \lora can also be seen as a form of regularization that reduces forgetting and can thereby actually improve performance \citep{JimenezYWYPPN24}.

\paragraph{Model Quantization and \qlora}
To reduce a model's memory footprint not only during training but also during inference, model quantization techniques have been proposed that reduce the precision of the model's weights and sometimes activations. In particular, representing the model's weight matrices using 4-, 3-, or even 2-bit precision rather than the standard (for LLMs) 16-bit half-precision representation can lead to significant memory savings \citep{ParkYV18,FrantarAHA22,LinTTYDH23}. However, quantization can also lead to a significant drop in performance, especially if applied after training. To mitigate this issue, \qlora \citep{DettmersPHZ23} proposes to quantize the weight matrices already during training, allowing the half-precision \lora adapters to learn to correct for the quantization errors, while still significantly reducing the memory footprint compared to standard \lora.

\section{Detailed Prompt descriptions}

\subsection{Root Cause Extraction}
\label{sec:app_prompt_extraction}
As described in \cref{sec:benchmark_construction}, we extract the root cause (RC) from the historic tickets using a strong LLM (\gptf) by concatenating the problem description, expert discussion, customer communication, and the final resolution with a chain-of-thought (CoT) prompt \citep{Wei0SBIXCLZ22} instructing the model to carefully analyze all provided information before generating a root cause description. We show the full prompt used to this end in \cref{fig:rc_extraction_prompt}.

\begin{figure}[t]
    \centering
    \begin{tcolorbox}[colback=white, colframe=black, arc=10pt, boxrule=0.8pt, left=10pt, right=10pt, top=10pt, bottom=10pt,width=0.85\linewidth]
    \ttfamily
    \small
    An issue with an industrial grade robot is reported in the context below. What is the root cause of the reported issue?
    
    \vspace{1em}

    Let's think step by step to answer this question. First, analyze each section in the context and systematically identify root causes and their relative probability. Remember that a section can have multiple root causes or no root causes at all. Finally, pick the root cause with the highest relative probability and respond with the root cause in JSON format with key as "Root Cause". If the root cause is unknown, respond "Unknown root cause" in JSON format.
    
    \vspace{1em}

    Context:

    \vspace{1em}

    Section for conversation between customer representative and engineers:
    \vspace{0.5em}
    \textcolor{red}{<CUSTOMER COMMUNICATION>}

    \vspace{1em}

    Section for conversation between engineers: 
    \vspace{0.5em}

    \textcolor{red}{<EXPERT DISCUSSION>}
    
    \vspace{1em}

    Section for analysis of issue:
    \vspace{0.5em}

    \textcolor{red}{<ISSUE RESOLTUION>}

    \end{tcolorbox}
    \caption{CoT prompt used for root cause extraction, where \textcolor{red}{<PLACEHOLDERS>} for data from the ticket are marked red.}
    \label{fig:rc_extraction_prompt}
    \vspace{-5mm}
\end{figure}

\subsection{Root Cuase Prediction}
\label{sec:app_prompt_prediction}
As described in \cref{sec:method}, we use a range of LLMs to predict the root causes of the tickets in \bench. To this end, we use zero-shot prompting with the problem description and our preprocessed logs (see \cref{sec:method}). We show the full prompt used for this task in \cref{fig:rc_prediction_prompt}.

\begin{figure}[t]
    \centering
    \begin{tcolorbox}[colback=white, colframe=black, arc=10pt, boxrule=0.8pt, left=10pt, right=10pt, top=10pt, bottom=10pt,width=0.85\linewidth]
    \ttfamily
    \small
    \textbf{instruction:} An issue with an industrial grade robot is reported in the input. Determine the root cause for the reported issue.
    
    \vspace{1em}

    \textbf{input:}

    Elog log message: 
    
    \textcolor{red}{<ELOG>}
    
    \vspace{0.5em}
    
    Error description: 
    
    \textcolor{red}{<PROBLEM DESCRIPTION>}
    
    \vspace{0.5em}
    
    Startup log message: 
    
    \textcolor{red}{<STARTUP>}
    
    \vspace{0.5em}
    
    Print spool log message: 
    
    \textcolor{red}{<PRINT SPOOL>}

    \vspace{1em}

    \textbf{output:} \textcolor{blue}{<REFERENCE ROOT CAUSE>}

    \end{tcolorbox}
    \caption{Prompt used for training and inference, where \textcolor{red}{<PLACEHOLDERS>} for data from the ticket are marked red and the \textcolor{blue}{<REFERENCE ROOT CAUSE>} is only provided during training. Depending on the model, we use the appropriate instruction template for both training and inference.}
    \label{fig:rc_prediction_prompt}
\end{figure}

\subsection{LLM-as-a-Judge: Similarity Score Prediction}
\label{sec:app_prompt_sim}
To assess the quality of generated root cause descriptions, we use a LLM-as-a-judge evaluation procedure \citep{ZhengC00WZL0LXZ23} where we ask a model to judge the similarity between the predicted and the reference root cause descriptions on a scale of $1$ to $10$. We show the full prompt used for this task in \cref{fig:similarity_score_prompt}.

\begin{figure}[t]
    \centering
    \begin{tcolorbox}[colback=white, colframe=black, arc=10pt, boxrule=0.8pt, left=10pt, right=10pt, top=10pt, bottom=10pt,width=0.85\linewidth]
    \ttfamily
    \small

    Please act as an impartial judge and evaluate the similarity of the two analyses provided below by two different AI assistants. Both were given the same data related to an issue in a robotics system and asked to identify the root cause. Your job is to evaluate and quantify the similarity between the two answers. Begin your evaluation by comparing the two answers and identifying key differences. Do not allow the length of the responses to influence your evaluation. Do not favor certain names of the assistant. Be as objective as possible. After providing your explanation, please rate the similarity on a scale of 1 to 10 by strictly following this format: "[[rating]]", for example: "Rating: [[5]]".

    \vspace{1em}

    [The Start of Analysis A] 

    $\;$\textcolor{red}{<PREDICTED ROOT CAUSE>}
    
    [The End of Analysis A]

    \vspace{1em}

    [The Start of Analysis B]
    
    $\;$\textcolor{red}{<REFERENCE ROOT CAUSE>}
    
    [The End of Analysis B]
    \end{tcolorbox}
    \caption{Prompt used for LLM-as-a-judge evaluation, where \textcolor{red}{<PLACEHOLDERS>} are replaced with the issue specific data.}
    \label{fig:similarity_score_prompt}
\end{figure}

\section{Experimental Setup}
\label{sec:app_setup}

\paragraph{Model Selection}
We select models based on three criteria: i) sufficient ($\geq 32$k tokens) context length, ii) good general reasoning capabilities, and iii) a permissive license. Based on these criteria, we choose \mixtral\footnote{https://huggingface.co/mistralai/Mixtral-8x7B-Instruct-v0.1} (\texttt{Mixtral-8x7B-Instruct-v0.1} under Apache-2.0 License \citealt{Mixtral2024}) and the smaller \mistral\footnote{https://huggingface.co/amazon/MistralLite} (\texttt{MistralLite} under Apache-2.0 License \citealt{MistralLite_2023}), which was specifically finetuned for long context tasks. As a reference frontier model, we consider \gptf (\texttt{gpt-4-32k-0613} \citealt{GPT4}).

\paragraph{Experimental Setup}
We use Axolotl \citep{Axolotl} with DeepSpeed \citep{RajbhandariRRH20,RasleyRRH20} for finetuning with AdamW ($\beta_1 = 0.9$ and $\beta_2=0.95$) \citep{LoshchilovH19} on $2$ to $8$ NVIDIA A100s. We train for 3 epochs at an effective batch size of $64$ for full finetuning and $16$ for \lora and \qlora using an initial learning rate of $10^{-5}$ and a cosine decay with a warm-up ratio of $10\%$. Unless indicated otherwise, we use rank $r=32$ for \lora and \qlora and NFloat4 + DQ (double quantization) for \qlora.

\paragraph{Computational Requirements}
We provide an overview of the runtimes required for our different training setups in \cref{tab:compute}.

\begin{table}[h]
    \caption{GPU (NVIDIA A100) hours required for different training setups.}
    \vspace{-2mm}
    \label{tab:compute}
    \centering
    \resizebox{0.55\linewidth}{!}{
    \begin{tabular}{ll
                    x{2}{0}
        }
        \toprule
        Model & Training Mode & {Training Time [h]} \\
        \midrule
        \multirow{3}{*}{\mistral} & FFT & 22.7\\
        & \lora & 20 \\
        & \qlora & 20 \\
        \cmidrule{1-1}
        \multirow{3}{*}{\mixtral} & FFT & 64\\        
        & \lora & 40 \\
        & \qlora & 40 \\
        \bottomrule
    \end{tabular}
    }
    \vspace{-4mm}
\end{table}

\section{Extended Experimental Evaluation}
\label{sec:app_experiments}

\paragraph{Similarity Score Callibration} We repeat the label extraction process described in \cref{sec:benchmark_construction} twice more for each ticket, sampling at a temperature of $t=0.5$, and compute similarity scores to the reference label, obtained with greedy decoding. We thus obtain an $MSS = 7.5$, for root causes extracted by the same strong model with access to the same information, yielding a reference for an excellent $MSS$.%

\vspace{-2mm}
\paragraph{Feature Importance Analysis}
To assess feature importance, we train \mistral on different feature subsets using \lora and report results in \cref{tab:feature_ablation}. We observe that the model's performance drops significantly when excluding the \texttt{elog} and Problem Description, indicating their importance. In contrast, removing the \texttt{startup} and \texttt{printspool} improves the model's performance, suggesting that while these features may be helpful to debug compilation issues, they are less important for reported issues and can even distract the model \citep{JimenezYWYPPN24}.
\begin{table}[h]
    \vspace{-4mm}
    \caption{Mean similarity score of \mistral with \lora training depending on features used.}
    \vspace{-3mm}
    \label{tab:feature_ablation}
    \centering
    \resizebox{0.45\linewidth}{!}{
    \begin{tabular}{l
                    x{2}{2}
        }
        \toprule
        Model & {$MSS$} \\
        \midrule
        All Features&  3.07  \\ 
        $\quad$ without \texttt{startup} & {$\;\,$}\textbf{3.17}\\
        $\quad$ without \texttt{printspool} & 3.11\\
        $\quad$ without \texttt{elog} & 2.85\\
        $\quad$ without Problem Description & 2.58\\
        \cmidrule(lr){1-2}
        $\quad$ without \texttt{printspool} and \texttt{startup}& {$\;\,$}\textbf{3.22} \\
        \bottomrule
    \end{tabular}
    }
    \vspace{-4mm}
\end{table}   

\begin{figure}[t]
    \centering
    \includegraphics[width=0.6\linewidth]{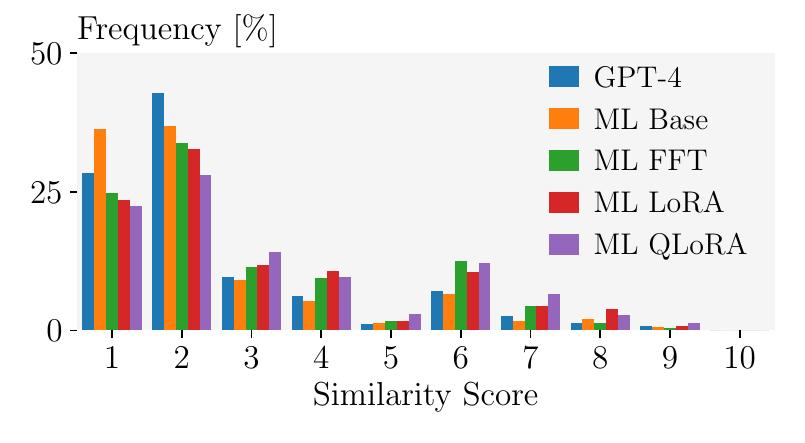}
    \vspace{-6mm}
    \caption{Similarity score distribution of \gptf and variants of \mistral (ML).}
    \label{fig:score_dist}
\end{figure}
\paragraph{Distribution of Similarity Scores}
Illustrating performance distributions in \cref{fig:score_dist}, we observe that distributions are similar across all models with better models having significantly fewer very low ($\leq2$) scores and a uniform increase in the frequency of all higher scores ($\geq3$).

\begin{figure}[t]
    \centering
    \includegraphics[width=0.6\linewidth]{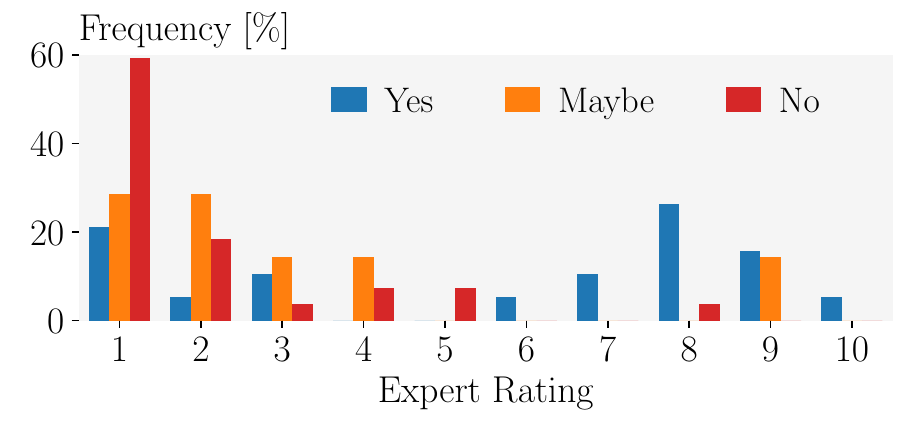}
    \vspace{-6mm}
    \caption{Human expert rating of fully finetuned \mistral depending on whether they considered the reference label to be correct (yes, maybe, or no).}
    \label{fig:rating_by_GT_correct}
\end{figure}

\paragraph{Model Helpfulness} Beyond the analysis in \cref{sec:experiments}, we visualize the expert rating of their highest-rated model depending on whether the experts consider our reference label to be correct in \cref{fig:rating_by_GT_correct}. We find that while average scores are higher when the reference label is considered correct, there is a significant number of examples, where the predicted RC is highly rated even when the reference label is considered incorrect. By manual inspection, we find that in these cases, identifying the root cause required understanding of the log files and was not possible from the expert discussion, communication with the customer, and issue resolution alone, which was used for our reference label extraction. Together, these results suggest that our models can provide valuable insights and help experts in diagnosing root causes of complex robotics systems issues.

\section{Human Expert Study}
\label{sec:app_human_study}

We illustrate the survey interface including all questions in \cref{fig:survey}. Unfortunately, we cannot provide examples of the analyzed data and identified RCs due to the proprietary nature of the data.

\begin{figure}
    \centering
    \includegraphics[width=0.432\linewidth]{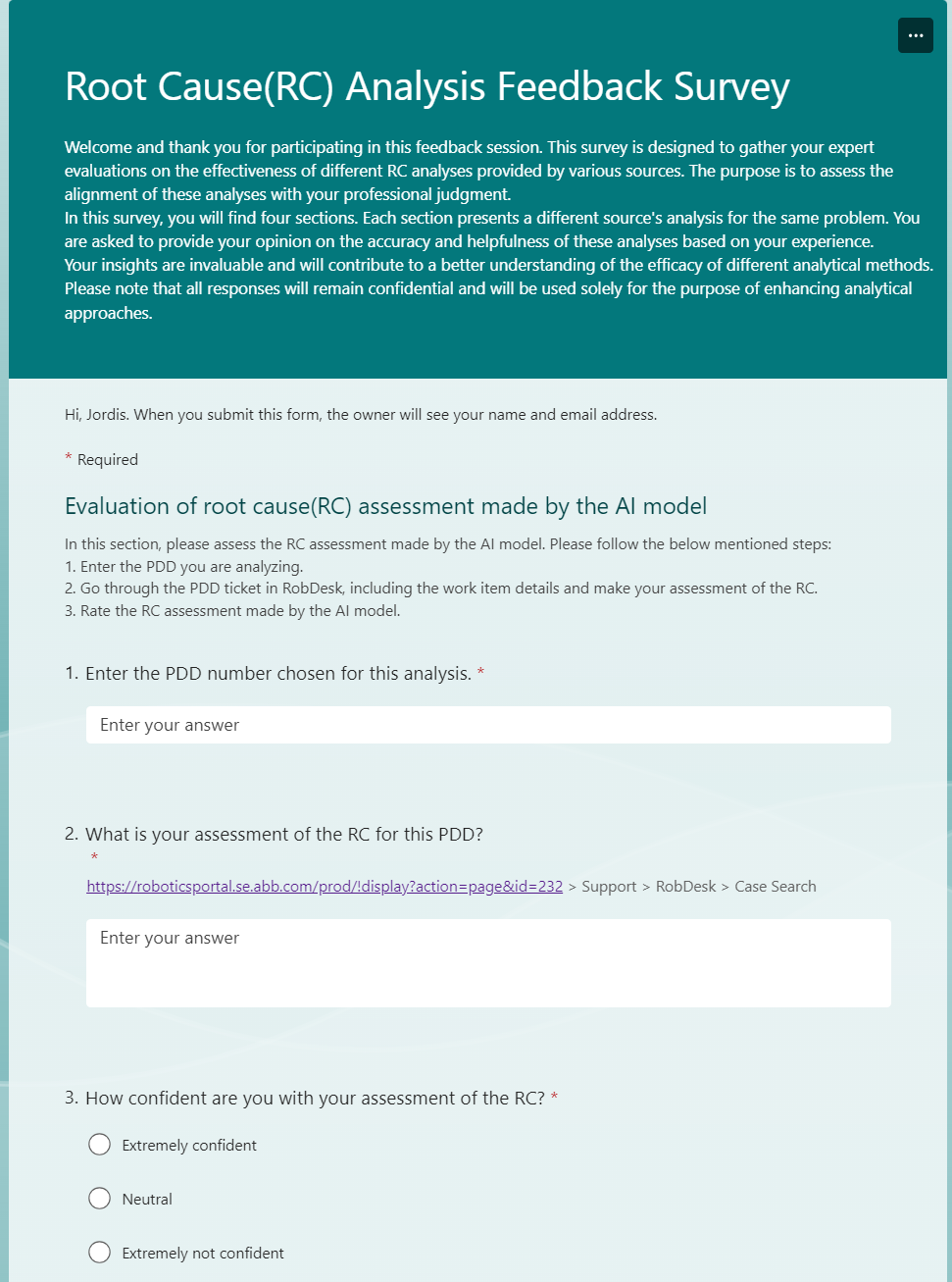}
    \hfil
    \includegraphics[width=0.450\linewidth]{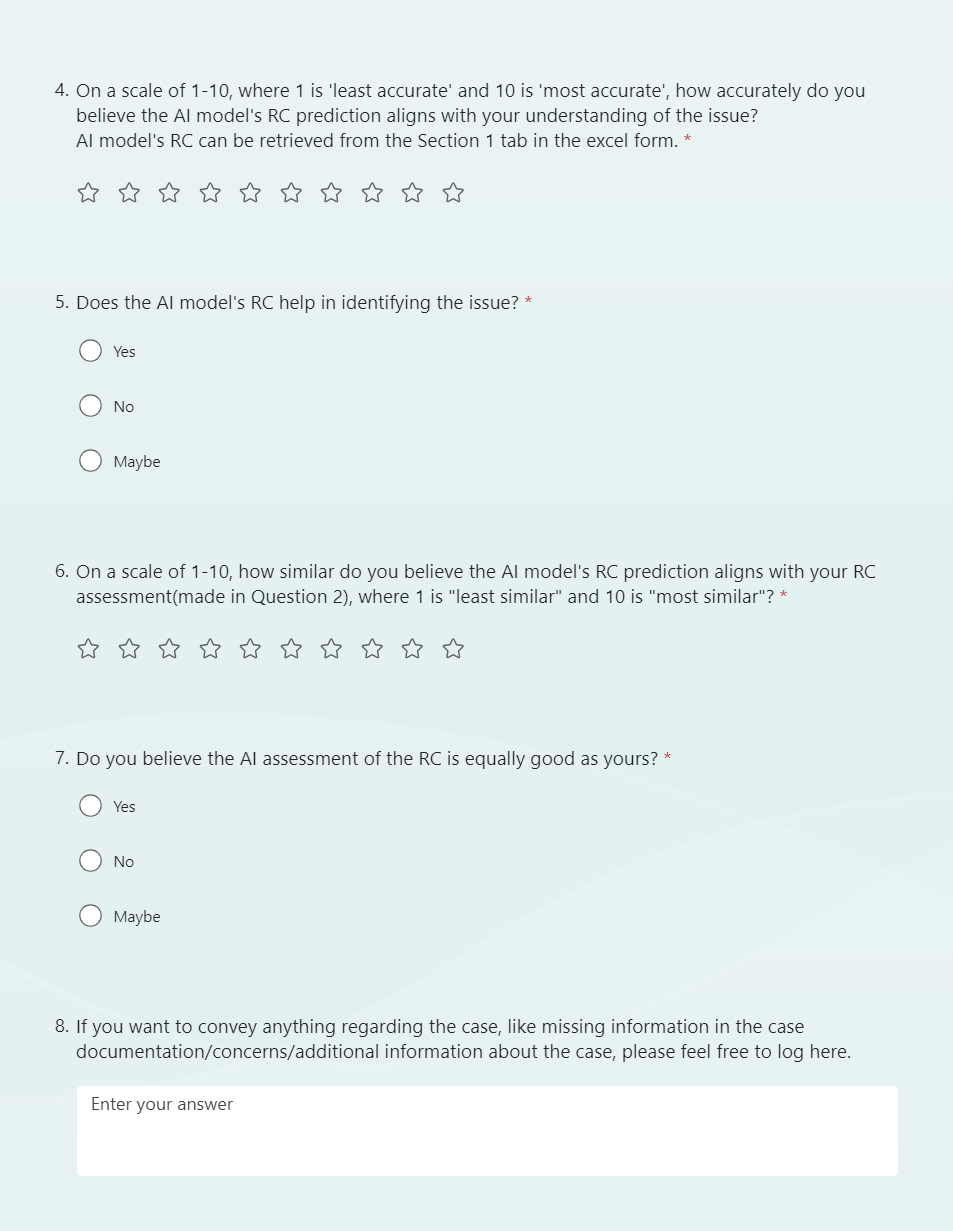}
    
    \vspace{3mm}

    \includegraphics[width=0.465\linewidth]{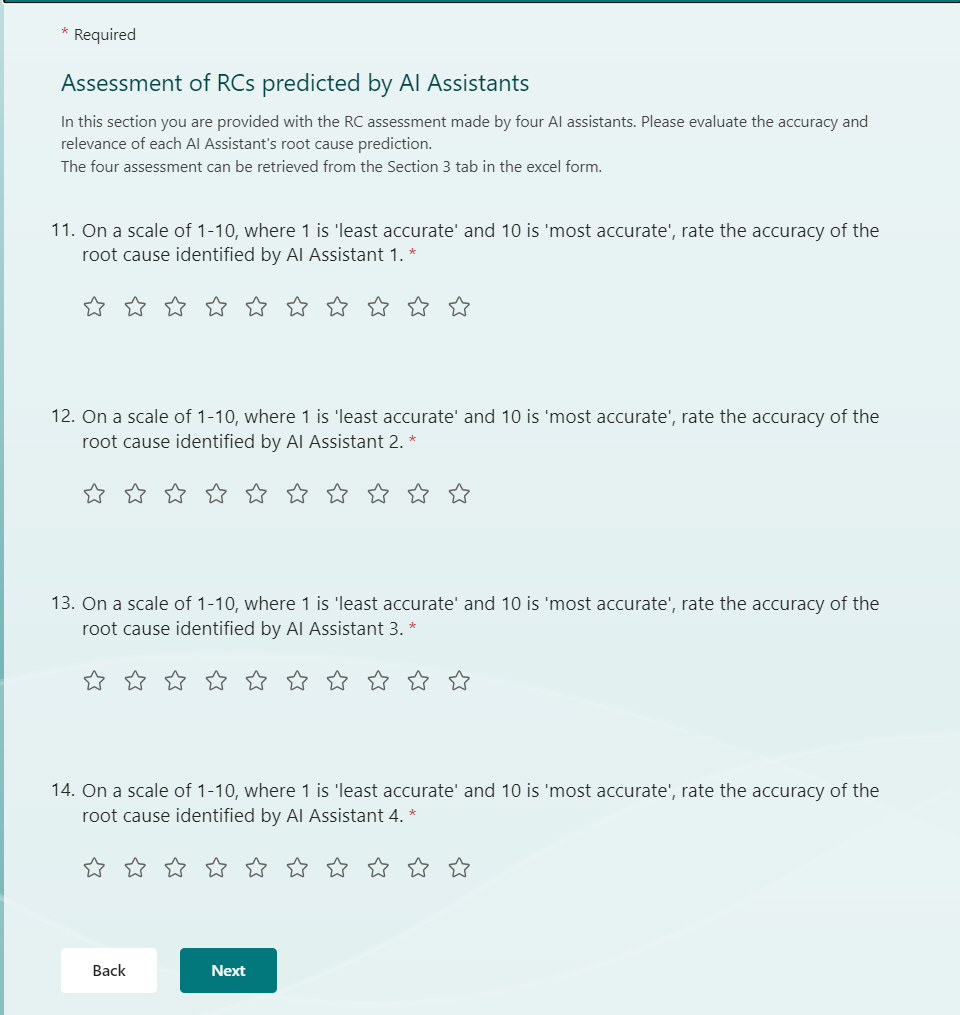}
    \hfil
    \includegraphics[width=0.430\linewidth]{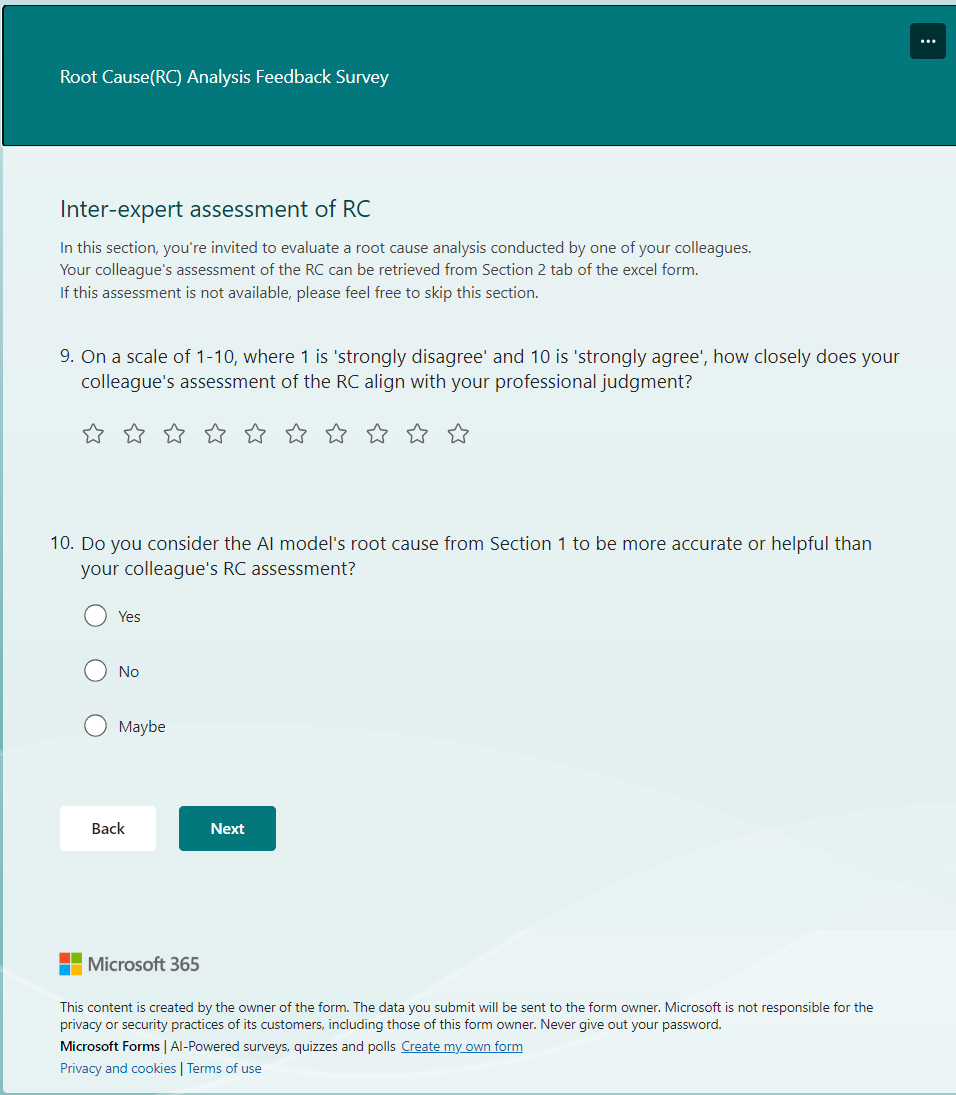}
    \caption{Human expert study questions and interface.}
    \label{fig:survey}
\end{figure}

\fi

\end{document}